\documentclass[11pt]{article}

\usepackage[final]{acl} 

\usepackage{times}
\usepackage{latexsym}
\usepackage[T1]{fontenc}
\usepackage[utf8]{inputenc}
\usepackage{microtype}
\usepackage{inconsolata}

\usepackage{graphicx}
\usepackage{tabularx}
\usepackage{booktabs}
\usepackage{soul}
\usepackage{siunitx}
\usepackage{algorithm}
\usepackage{algorithmic}
\usepackage[british]{babel}
\usepackage{enumitem}
\usepackage[normalem]{ulem}
\usepackage{cuted}

\usepackage{framed}
\definecolor{addgreen}{RGB}{255,255,255}
\definecolor{delgrey}{RGB}{125,125,125}
\newenvironment{added}{}{}
\sethlcolor{addgreen}
\newcommand{\ins}[1]{#1}
\newcommand{\del}[1]{#1}

\hypersetup{
    colorlinks=true,
    linkcolor=blue,
    filecolor=magenta,
    urlcolor=blue,
    citecolor=blue,
}

\title{AutoSpecNER: A Fine-Grained Named Entity Recognition Dataset for Vehicle Specification Extraction}

  \author{
    \textbf{Jordan Lee}\textnormal{\textsuperscript{1,2,*},}
    \textbf{Filippos Ventirozos}\textnormal{\textsuperscript{1,2},}
    \textbf{Abdirahman Abdullahm}\textnormal{\textsuperscript{1},}
    \textbf{Ioanna Nteka}\textnormal{\textsuperscript{2},} \\
    \textbf{Peter Appleby}\textnormal{\textsuperscript{2}},
    \textbf{Matthew Shardlow}\textnormal{\textsuperscript{1}} \\
    \textsuperscript{1}Department of Computing and Mathematics, Manchester Metropolitan University, UK \\
    \textsuperscript{2}Autotrader Research Group, Autotrader UK \\
    \texttt{\{f.ventirozos,m.shardlow\}@mmu.ac.uk} \\
    {\small \textsuperscript{*}Work conducted during an internship at Autotrader UK.}
  }

\begin{document}
\maketitle

\begin{abstract}
Vehicle advertisements contain rich specification information, but automotive NER resources remain limited. We introduce \textbf{AutoSpecNER}, an expert-annotated dataset for fine-grained entity recognition in vehicle listings. The dataset includes 659 advertisements from a popular car-selling website, with over 10,000 entities annotated across 15 categories, including \texttt{MODEL}, \texttt{ENGINE\_SPEC}, and \texttt{BATTERY\_CAPACITY}. Annotation quality was validated through inter-annotator agreement, achieving an average score of 91.5\%. We benchmark rule-based extraction, fine-tuned transformer encoders, and large language models. DeBERTa achieves the best performance with a 90\% micro-F1 score, outperforming the rule-based baseline (43\%) and the strongest large language model (77.8\%).
\end{abstract}


\section{Introduction}

The automotive industry generates vast amounts of unstructured text through online vehicle advertisements. These advertisements contain valuable specification information embedded in free-form descriptions, but extracting this structured data manually is impractical at scale. This challenge is compounded by the recent introduction of AI-generated advertisement content, which can contain hallucinations---factually incorrect information that may mislead consumers.

Named entity recognition (NER) offers a solution by automatically identifying and extracting specific information spans from text. While NER has been successfully applied to various domains including biomedical texts \citep{Majid:2024}, news articles \citep{Sang2003}, and social media \citep{Derczynski2017}, the automotive advertisement domain presents unique challenges:

\begin{itemize}
\item \textbf{Domain-specific terminology}: Technical specifications like ``2.0L TDI'', ``DSG transmission'', or ``Santorini [black colour]'' require specialised understanding.
\item \textbf{Fine-grained distinctions}: Differentiating between similar concepts (e.g., exterior vs. interior colour, battery capacity vs. range).
\item \textbf{Multi-word entities}: Complex specifications often span multiple tokens (e.g., "18 minutes with 350kW charger").
\item \textbf{Mixed content sources}: Advertisements include both user-generated content with typos and informal language, and AI-generated content with potential hallucinations.
\end{itemize}




This paper presents two main contributions. First, we introduce Automotive Specification NER (\textbf{AutoSpecNER}), a new, publicly available dataset\footnote{Available at \\\href{https://github.com/FilipposVentirozos/AutoSpecNER}{github.com/FilipposVentirozos/AutoSpecNER}} for fine-grained NER tailored to the automotive domain. The dataset contains 659 vehicle advertisements annotated with 15 entity types critical for vehicle identification and comparison.

Second, we provide a comprehensive benchmark evaluation on AutoSpecNER. We compare the performance of three diverse approaches: a rules-based system, transformer-based encoder models, and large language models (LLMs) prompted using few-shot and self-verification techniques \citep{wang2023gpt}. Our analysis confirms that while NER is a viable technique for this task, the choice of model has a profound impact on performance, with fine-tuned encoders demonstrating superior capabilities.

\section{Related Work}

\subsection{Domain-Specific NER and Fine-Grained Entity Recognition}
Standard NER benchmarks like CoNLL-2003 \citep{Sang2003} focus on coarse-grained entities (person, location, organisation) inadequate for technical domains. Fine-grained entity recognition \citep{ling2012fine} requires distinguishing between closely related entity types—a challenge particularly acute in automotive contexts where hierarchical relationships exist between entities (e.g., ``2024 Ford F-150 Limited'' contains YEAR, MAKE, MODEL, and TRIM entities).

Recent work has explored product and attribute extraction \citep{putthividhya2011bootstrapped,chen2023does}, demonstrating unique challenges in technical NER: domain-specific abbreviations, overlapping entity boundaries, and hierarchical entity relationships. Fine-grained annotation schemas have proven essential for capturing technical specifications in industrial domains \citep{bikaun2024maintie}, yet automotive-specific resources remain limited.

\subsection{Automotive NER Research}
The automotive domain has received minimal attention in NER research. \citet{hu2024named} addressed NER for automotive accessories in Chinese, while \del{recent }work by \citet{ventirozos2024shifting} introduces the Auto-AdvER approach for English vehicle advertisements to understand the condition, historic claims and sales options offered. \del{Recent datasets}\ins{Datasets} like FindVehicle \citep{guan2024findvehicle} target vehicle retrieval with entity types including vehicle colour, brand, model, and location, but lack the granularity needed for technical specification extraction.

\citet{park2023admit} introduced ADMit, combining adversarial training and multi-task learning for automotive NER for \del{a} Korean and English. Their work addresses domain adaptation between general and automotive-specific terminology but focuses on FAQ systems rather than technical specifications. Our work differs by targeting fine-grained vehicle identifiables critical for specification verification and hallucination detection.

\subsection{Neural Approaches and Domain Adaptation}
Transformer-based models have become standard for NER, with BERT \citep{devlin-etal-2019-bert} and its variants achieving strong baseline performance. Domain-specific pre-training significantly improves technical entity recognition, as demonstrated in manufacturing domains where morphological patterns guide entity recognition \citep{li2024corpus}.

Few-shot NER methods combining knowledge graphs and contrastive learning show promise for low-resource domains \citep{zhang2024kcl}, addressing the scarcity of labelled automotive data. While LLMs demonstrate competitive performance through prompting \citep{wang2023gpt}, recent studies \citep{naguib-etal-2024-shot} show that smaller, specialised models often outperform LLMs in low-resource technical domains, making them more practical for deployment in automotive applications.

\section{The AutoSpecNER Dataset}

\subsection{Data Collection and Composition}

We were provided 659 vehicle advertisements from one of the UK's biggest online vehicle-selling websites. The dataset comprises two distinct sources:

\textbf{User-generated advertisements (350)}: 
The dataset was a sample of an equal distribution of ads written by individual sellers (stratified across different UK counties) and dealerships (stratified across the largest dealerships in the UK). The advertisements contain natural language variation, informal descriptions, typographical errors, and inconsistent formatting. 

Example: \textit{``\uwave{lovley} ford focus 1.8 \uwave{diesal}, 2015 plate, full mot till next yr, grey \uwave{metalic} paint''}.

\textbf{AI-generated advertisements (309)}: Created\footnote{These were generated by the team of the aforementioned UK website company. These AI generated descriptions were shown to the users and the users decided whether they want to keep them or edit/delete them.} using Google Gemini\footnote{gemini-2.0-flash-001} and Meta's LLaMA3\footnote{llama-3.1-8b-instruct} by providing in the prompt the vehicle specifications. These are grammatically correct and well-structured but may contain hallucinations. For example in this advertisement for an Audi RSQ8, where the generated advert incorrectly refers to the vehicle as an Audi RS6:

\begin{quote}
    ``The Audi \uwave{RS6} is a high-performance car that boasts a powerful 4.0-litre V8 engine. This petrol engine is paired with an automatic transmission...''
\end{quote}

Other hallucinations exist which are more subtle than this such as many adverts have additional information inserted not present in the specifications, but could be true. An example of this is in the following advert in which a Volkswagen Polo Match is referred to as a Volkswagen Polo EVO Match:

\begin{quote}
    ``With only 21,029 miles on the clock, this 2021 Volkswagen Polo \uwave{EVO} Match is manufacturer approved...''
\end{quote}

This dual-source approach enables investigation of how NER models handle both human errors (typos, informality) and AI errors (hallucinations, specification confusion). Crucially, each advertisement is accompanied by structured meta-data (a fact table) that lists the vehicle's actual specifications. This is a vital characteristic, as it allows for the verification of extracted entities and forms the basis for potential error/hallucination detection systems.


\subsection{Corpus Annotation and Inter-Annotator Agreement}
\label{sec:annotation}

To ensure the reliability and interpretability of our proposed 15-label schema, we undertook a multi-stage annotation process. The process was iterative, involving an initial schema refinement phase to produce clear guidelines\footnote{The annotation guidelines can be found at \href{https://github.com/FilipposVentirozos/AutoSpecNER}{github.com/FilipposVentirozos/AutoSpecNER}.}, followed by a final validation phase to confirm their efficacy.

Our initial phase involved two annotators independently labelling a pilot set of 50 advertisements, balanced between user- and AI-generated content. While this yielded a promising micro F1-score of 0.81, a qualitative analysis of the disagreements revealed systematic ambiguities. This analysis was crucial for developing a set of explicit annotation principles.

The key principles that emerged from this refinement process were:

\begin{itemize}[leftmargin=10pt]

    \item \textbf{Prioritise Specificity over General Description:} Labels are reserved for specific, quantifiable details, not general descriptive statements. For instance, the phrase ``low CO2 emissions'', while describing an engine's characteristic, is not annotated as \textit{ENGINE\_SPEC} because it lacks a specific value.

    \item \textbf{Maintain Entity Separation for Adjacent, Distinct Items:} When multiple entities of the same type appear adjacently but refer to distinct details, they must be annotated as separate entities. For example, in the text ``...TDCI 1.6L ECOnetic...'', each component (``TDCI'', ``1.6L'', ``ECOnetic'') is labelled as a separate \textit{ENGINE\_SPEC} entity.

    \item \textbf{Disambiguate Overlapping Concepts:} Clear distinctions were established to prevent confusion. The \textit{MAKE} label, for example, is restricted to the primary vehicle manufacturer; in `...a Ford Focus with a Mercedes...engine`, only ``Ford'' is labelled as \textit{MAKE}. Similarly, composite names like ``e-SKYACTIV G Centre-Line'' are split into ``e-SKYACTIV G'' (\textit{ENGINE\_SPEC}) and ``Centre-Line'' (\textit{TRIM}).

    \item \textbf{Enforce Strict Contextual Boundaries:} Annotators were instructed to capture the full, self-contained descriptive phrase. For \textit{BATTERY\_RANGE}, the entire phrase ``maximum range of 280 miles when new'' is captured, preserving the qualifying context.

    \item \textbf{Exclude Speculative and Ancillary Information:} The guidelines were refined to ensure that only the direct attribute of the advertised vehicle is annotated. In cases where an advert mentions other available models (e.g., ``The Golf is also available as a Station Wagon (Estate)''), only the body type of the primary advertised vehicle (e.g., ``Hatchback'') is to be labelled.

    \item \textbf{Annotate the Text, Not the Structured Specification:} 
    Annotations should reflect the entity span as it appears in the advertisement, even when it differs from the accompanying structured specification data or appears factually incorrect. For example, if the advert describes the trim as ``220d Luxury'', annotators label the span in the text rather than normalising it to the official specification value, ``Luxury''. This ensures that models learn to extract the claims made in the free-text advert, which can later be compared against structured specifications to detect inconsistencies or hallucinations.
\end{itemize}

\begin{added}
\paragraph{Annotation Effort and Procedure}
All annotation was carried out in Prodigy\footnote{\href{https://prodi.gy}{prodi.gy}} by members of the author team rather than by crowdworkers. Each advertisement was annotated in two passes: a first full labelling pass, followed by a second pass over the same document to recover missed spans, since omission---rather than mislabelling---was the dominant error observed during the pilot. Effort scaled with specification density rather than document length: the longer but predictably structured AI-generated adverts were quicker to annotate than terse user-written ones.
\end{added}

\subsection{Annotation Schema Overview}
The final schema comprises 15 entity labels, organised into four groups and defined with representative examples in Table~\ref{tab:schema}.

\begin{table}[t]
\centering
\footnotesize
\setlength{\tabcolsep}{4pt}
\renewcommand{\arraystretch}{1.15}
\begin{tabularx}{\linewidth}{@{}l>{\raggedright\arraybackslash}X>{\raggedright\arraybackslash}Xcc@{}}
\toprule
\textbf{Label} & \textbf{Definition} & \textbf{Example} & \ins{\textbf{IAA}} & \ins{\textbf{\mbox{$n$}}} \\
\midrule
\multicolumn{5}{@{}l}{\textit{Vehicle identity}}\\
\texttt{MAKE}             & Vehicle manufacturer & ``CUPRA'' & \ins{.968} & \ins{114} \\
\texttt{MODEL}            & Specific model name & ``Focus'' & \ins{.962} & \ins{178} \\
\texttt{TRIM}             & Trim level / variant of a model & ``220d Luxury'' & \ins{.868} & \ins{129} \\
\texttt{YEAR}             & Registration year & ``2021'' & \ins{.967} & \ins{61} \\
\addlinespace
\multicolumn{5}{@{}l}{\textit{Powertrain}}\\
\texttt{ENGINE\_SPEC}     & Engine detail: volume, cylinders, or technology code & ``TDCI'' & \ins{.912} & \ins{104} \\
\texttt{FUEL\_TYPE}       & Fuel powering the vehicle & ``petrol'' & \ins{.936} & \ins{64} \\
\texttt{TRANSMISSION}     & Gear system type & ``automatic'' & \ins{.732} & \ins{13} \\
\addlinespace
\multicolumn{5}{@{}l}{\textit{Body \& appearance}}\\
\texttt{BODY\_TYPE}       & Shape/style of the chassis & ``Panel Van'' & \ins{.867} & \ins{20} \\
\texttt{EXTERIOR\_COLOUR} & Exterior paint colour, incl. descriptors & ``Black sapphire metallic'' & \ins{.866} & \ins{25} \\
\texttt{INTERIOR\_COLOUR} & Colour/material of interior features & \quad``black leather seats'' & \ins{.639} & \ins{16} \\
\texttt{NO\_SEATS}        & Total seating capacity & ``5 seats'' & \ins{.769} & \ins{4} \\
\texttt{BOOT\_SIZE}       & Boot storage capacity (litres) & ``6100 litres'' & \ins{1.00} & \ins{1} \\
\addlinespace
\multicolumn{5}{@{}l}{\textit{Electric vehicle}}\\
\texttt{BATTERY\_CAPACITY}& Battery capacity (kWh) & ``68 kWh'' & \ins{.961} & \ins{17} \\
\texttt{BATTERY\_RANGE}   & Travel distance, incl. qualifiers & ``max range of 280 miles when new'' & \ins{1.00} & \ins{10} \\
\texttt{RECHARGE\_TIME}   & Recharge time, incl. charger/level context & ``empty to 80\% in as little as 53 minutes'' & \ins{.693} & \ins{17} \\
\bottomrule
\end{tabularx}
\caption{The 15-label \textbf{AutoSpecNER} annotation schema, grouped by category with definitions and representative examples. \ins{IAA is the mean of the three pairwise inter-annotator F1 scores from the agreement study of \mbox{Section~\ref{sec:iaa-final}}; \mbox{$n$} is the mean number of instances of the label annotated per annotator in that study.}}
\label{tab:schema}
\end{table}

\subsubsection{Final Validation and Agreement}
\label{sec:iaa-final}

Following the guideline refinement, a final inter-annotator agreement study was conducted. This involved three annotators (the authors of this paper) of mixed nationalities with English as either their first or second professional language and varying levels of prior exposure to NLP annotation tasks, who labelled a new, larger random sample of 100 advertisements (50 user-generated and 50 AI-generated). Using the finalised guidelines, this validation achieved a strict-matching average micro F1-score\footnote{In NER in most cases Kappa is equivalent to F1, hence, we just measure F1 \citep{richie-etal-2022-inter,hripcsak2005agreement}} of \textbf{91.5\%} across all three annotators (standard deviation: \textbf{3.2\%}). \del{Precision consistently exceeded recall by approximately 2 percentage points, indicating strong agreement on positive entity identification while annotators exhibited greater conservatism when entity boundaries or labels were ambiguous.}

\begin{added}
Because agreement is measured between annotators rather than against a reference, precision and recall are interchangeable under exchange of the two annotators, and we therefore report only F1. Annotation density was nonetheless consistent: the three annotators marked 765, 788 and 768 entities respectively, a spread of under 3\%.

Aggregate agreement conceals variation across the schema; per-label figures are given in Table~\ref{tab:schema} and, with support and mean entity length, in Table~\ref{tab:iaa_full}. Agreement is near-ceiling for categories with canonical surface forms---\texttt{MAKE} (.968), \texttt{YEAR} (.967), \texttt{MODEL} (.962)---but lower for \texttt{INTERIOR\_COLOUR} (.639), \texttt{RECHARGE\_TIME} (.693) and \texttt{TRANSMISSION} (.732), the categories the annotators found hardest, though the small support of these labels in the agreement sample means their estimates should be read with caution. \texttt{INTERIOR\_COLOUR} requires a judgement of significance the guidelines could not fully operationalise (annotators divided over intermediate cases such as trim inserts and door furniture); \texttt{RECHARGE\_TIME} is by far the longest entity type, averaging 65.5 characters, and its disagreements concern how much qualifying context to absorb into the span; \texttt{TRANSMISSION} shows the same boundary problem, as gear-system descriptions are often embedded in longer powertrain phrases.

Decomposing the disagreements (Table~\ref{tab:iaa_anatomy}), 91.7\% of annotated spans match exactly in both offsets and label, 3.9\% overlap but disagree on boundaries, 3.3\% are marked by one annotator of a pair only, and 1.1\% assign different labels to an identical span: disagreement is chiefly about span delimitation and omission rather than category semantics. The few label confusions cluster around compound designations in which one surface string encodes several facts---``MHEV'' (\texttt{ENGINE\_SPEC} vs.\ \texttt{FUEL\_TYPE}) or BMW's ``520d'' (\texttt{ENGINE\_SPEC} vs.\ \texttt{MODEL})---where a single-label schema forces a choice that we fixed by convention in the guidelines. Omissions concentrate in the two densest categories, \texttt{TRIM} and \texttt{ENGINE\_SPEC}.
\end{added}

\begin{center}
\begin{minipage}{\columnwidth}
\centering
\small
\begin{tabular}{@{}lccc@{}}
\toprule
\textbf{Label} & \textbf{Mean F1} & \textbf{Support} & \textbf{Len.} \\
\midrule
INTERIOR\_COLOUR  & 0.639 & 16.0  & 19.3 \\
RECHARGE\_TIME    & 0.693 & 17.3  & 65.5 \\
TRANSMISSION      & 0.732 & 13.3  & 7.1 \\
NO\_SEATS         & 0.769 & 4.3   & 7.1 \\
EXTERIOR\_COLOUR  & 0.866 & 24.7  & 15.2 \\
BODY\_TYPE        & 0.867 & 20.3  & 6.7 \\
TRIM              & 0.868 & 128.7 & 8.9 \\
ENGINE\_SPEC      & 0.912 & 104.3 & 4.5 \\
FUEL\_TYPE        & 0.936 & 64.0  & 8.0 \\
BATTERY\_CAPACITY & 0.961 & 16.7  & 6.0 \\
MODEL             & 0.962 & 177.7 & 4.7 \\
YEAR              & 0.967 & 61.3  & 4.1 \\
MAKE              & 0.968 & 114.0 & 4.9 \\
BATTERY\_RANGE    & 1.000 & 10.0  & 35.0 \\
BOOT\_SIZE        & 1.000 & 1.0   & 11.0 \\
\bottomrule
\end{tabular}
\captionof{table}{Per-label inter-annotator agreement, ascending. Support is the mean number of annotated instances per annotator; Len.\ is the mean entity length in characters.}
\label{tab:iaa_full}
\end{minipage}
\end{center}

\begin{center}
\begin{minipage}{\columnwidth}
\centering
\small
\begin{tabular}{@{}lc@{}}
\toprule
\textbf{Disagreement type} & \textbf{Share} \\
\midrule
Exact match (offsets and label) & 91.7\% \\
Overlapping span, different boundaries & 3.9\% \\
Annotated by one annotator only & 3.3\% \\
Identical span, different label & 1.1\% \\
\bottomrule
\end{tabular}
\captionof{table}{Anatomy of annotator disagreement, pooled over the three annotator pairs.}
\label{tab:iaa_anatomy}
\end{minipage}
\end{center}

\subsection{Dataset Statistics}
\begin{figure}[h]
    \centering
    \includegraphics[width=0.99\linewidth]{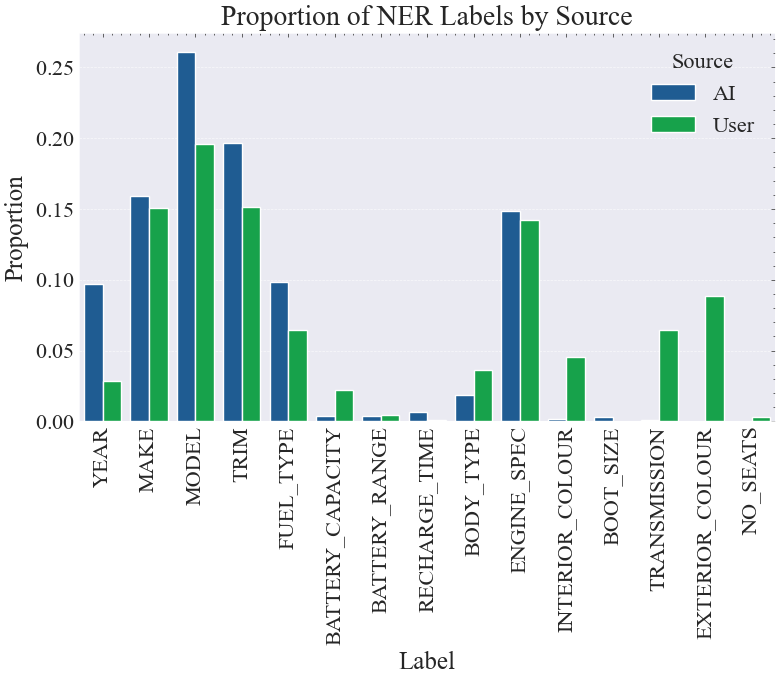}
    \caption{Proportion of labels from each source within the dataset. In dark blue are \del{the} the text advertisements generated by either Gemini or Llama and in green are the ones written by users.}
    \label{fig:label_proportion}
\end{figure}

The dataset contains a total of 11,117 labelled entities overall. Entity frequencies approximate a power law distribution, from MODEL (2,426 instances) to NO\_SEATS (10 instances). This reflects natural occurrence patterns - every advertisement mentions the model, but seat count is specified only when notable. One can look into Figure~\ref{fig:label_proportion} to view the label distributions for the user and the AI advertisement generated text respectively. \del{In total, the dataset contains 44 different brands, 237 unique models and the vehicles range from older (first registered in 2006) to newest (this year 2025).} \ins{In total, the dataset covers 44 brands and 237 unique models, with vehicles first registered between 2006 and 2025.}

\section{Experiments}

To prepare the annotated corpus for training, we performed two key steps: pre-processing the span-level annotations into a token-level format and partitioning the corpus into training (\textasciitilde70\%), validation (\textasciitilde15\%), and testing (\textasciitilde15\%) sets, the distribution of the entities can be found in Table~\ref{tab:dataset_splits}.


\newcommand{\datasetSplitTable}{%
\begin{center}
\begin{minipage}{\columnwidth}
\centering

\scalebox{0.9}{
\sisetup{group-separator={,}} 
\begin{tabular}{@{}l S[table-format=4.0] S[table-format=3.0] S[table-format=3.0] S[table-format=5.0]@{}}
\toprule
\textbf{Label} & {\textbf{Train}} & {\textbf{Validation}} & {\textbf{Test}} \\
\midrule
ENGINE\_SPEC & 1694 & 283 & 468 \\
MODEL & 1630 & 347 & 449  \\
TRIM & 1399 & 285 & 349  \\
MAKE & 838 & 172 & 219  \\
FUEL\_TYPE & 483 & 70 & 120  \\
YEAR & 353 & 75 & 111  \\
EXTERIOR\_COLOUR & 297 & 58 & 125 \\
RECHARGE\_TIME\ins{\mbox{\textsuperscript{\mbox{$\dagger$}}}} & 260 & 78 & 10 \\
INTERIOR\_COLOUR & 178 & 16 & 107 \\
BODY\_TYPE & 116 & 32 & 33  \\
TRANSMISSION\ins{\mbox{\textsuperscript{\mbox{$\dagger$}}}} & 114 & 9 & 16 \\
BATTERY\_CAPACITY & 86 & 44 & 32 \\
BATTERY\_RANGE\ins{\mbox{\textsuperscript{\mbox{$\dagger$}}}} & 71 & 41 & 3 \\
BOOT\_SIZE\ins{\mbox{\textsuperscript{\mbox{$\dagger$}}}} & 16 & 8 & 12 \\
NO\_SEATS\ins{\mbox{\textsuperscript{\mbox{$\dagger$}}}} & 4 & 0 & 6  \\
\midrule
\textbf{Overall} & \textbf{7539} & \textbf{1518} & \textbf{2060}  \\
\bottomrule
\end{tabular}
}
\captionof{table}{Distribution of entity counts across the final training, validation, and testing sets\ins{. \mbox{$\dagger$} marks labels with fewer than 20 test instances: per-label scores for these are point estimates over very small samples and are not reliable for comparing models (\mbox{Section~\ref{sec:limitations}}).}}
\label{tab:dataset_splits}
\end{minipage}
\end{center}
}


The three different ML approaches operate on different units: sub-word tokens for the encoders, whitespace tokens for the rule-based matcher, and LLM-generated free-form text. We therefore evaluate every system on a single, tokeniser-independent character-level IOB2 representation. Specifically, for each advert, the gold character-offset spans are expanded into a per-character IOB2 sequence in which every character carries exactly one tag. Within an entity span, the first character is tagged \texttt{B-} and each subsequent character—including whitespace within a multi-word entity—is tagged \texttt{I-}; every character that falls outside any entity span is tagged \texttt{O} (``Outside''). Each system's predictions are mapped onto the same per-character IOB2 representation, so all systems are scored on identical units; we then reconstruct entities and report entity-level precision, recall, and micro-F1 with \texttt{seqeval}~\cite{seqeval}.

\datasetSplitTable

\subsection{Models}
We evaluated three classes of models on the AutoSpecNER dataset to establish a robust performance benchmark.

\subsubsection{Rules-Based Approach} 

To establish a performance baseline and explore a non-AI alternative, we developed a hybrid rules-based system for entity extraction. This approach was designed to be computationally inexpensive at inference time and serves as a benchmark against which our neural models are compared. The system combines two core methodologies: taxonomy-based matching and regular expressions.

\paragraph{Methodology}
Our system employs a two-pronged strategy, applying different techniques based on the nature of the target entity.

\paragraph{Taxonomy-Based Matching}
For eight of the core vehicle attributes (\texttt{MAKE}, \texttt{MODEL}, \texttt{TRIM}, \texttt{FUEL\_TYPE}, \texttt{BODY\_TYPE}, \texttt{TRANSMISSION}, \texttt{INTERIOR\_COLOUR}, and \texttt{EXTERIOR\_COLOUR}), we leveraged an external taxonomy provided by the company that provided the data. This taxonomy acts as a gazetteer of known values for each entity. The system iterates through the advertisement text, comparing text spans to the values in the gazetteer. We evaluated several matching strategies:
\begin{itemize}
    \item \textbf{Strict Matching:} Requires an exact, case-insensitive string match.
    \item \textbf{Partial Matching:} To account for minor variations and typographical errors, we implemented partial matching using a normalised Levenshtein distance function. Predictions were made only if the similarity score exceeded a pre-determined threshold.
    \item \textbf{Heuristic Filtering:} To improve precision, we experimented with two disambiguation heuristics: (1) a \textit{most-common-value} filter, which retains only the most frequently matched entity for labels expected to appear once (e.g., \texttt{MAKE}); and (2) \textit{hierarchical filtering}, which uses known make-model-trim relationships from the taxonomy to prune invalid predictions (e.g., removing a predicted trim that does not belong to the predicted model).
\end{itemize}

\paragraph{Regular Expressions}
For entities with highly predictable syntactic patterns that were not present in the taxonomy, we employed regular expressions. This approach was used for three specific labels: \texttt{BATTERY\_CAPACITY}, \del{\mbox{NUMBER\_OF\_SEATS}}\ins{\mbox{\texttt{NO\_SEATS}}}, and \texttt{YEAR}. The patterns, detailed in Table~\ref{tab:regex_patterns}, were designed to be robust enough to capture common formulations of these attributes.

\begin{strip}
\centering
\small

\begin{tabular}{@{}lll@{}}
\toprule
\textbf{Label} & \textbf{Pattern} & \textbf{Description} \\
\midrule
BATTERY\_CAPACITY & \texttt{\textbackslash b(\textbackslash w+)\textbackslash s+kWh\textbackslash b} & Matches numeric values followed by "kWh". \\
NO\_SEATS & \texttt{(\textbackslash d+)\textbackslash s*\textbackslash w*seat\textbackslash w*} & Matches seat counts (e.g., "5 seats"). \\
YEAR & \texttt{\textbackslash b(?:19{[}0-9{]}...)} & Matches four-digit years from 1900-2030. \\
\bottomrule
\end{tabular}
\captionof{table}{Regular expression patterns used for specific entity extraction.}
\label{tab:regex_patterns}
\end{strip}

\subsubsection{Transformer Encoders} We fine-tuned four widely-used encoder models: BERT-base-cased \citep{devlin-etal-2019-bert}, RoBERTa-base \citep{Liu2019Roberta}, ModernBERT-base \citep {warner-etal-2025-smarter} and DeBERTa-v3-base \citep{he2023deberta}. To perform the training and inference, we utilised the HuggingFace library \citep{wolf2020huggingfacestransformersstateoftheartnatural} for token classification.

We perform grid search over learning rates $\{5\times10^{-5}, 3\times10^{-5}\}$, batch sizes $\{4, 8, 16\}$, and weight decay values $\{0.0, 0.01, 0.1\}$ to identify optimal hyperparameters for each encoder architecture. We train all models for a maximum of 50 epochs with early stopping patience of 3 epochs, monitoring the weighted F1 score on the validation set. The best configuration across all models uses a learning rate $2\times10^{-5}$, batch size 8, weight decay 0.01, with a warmup of 100 steps and a maximum sequence length of 512 tokens used in all experiments.

\subsubsection{Large Language Models}

To provide a contemporary comparison to encoder-based models, we evaluated four open sourced LLMs for the NER task: Qwen3~\citep{qwen3} Mistral-7B \citep{Mistral}, LLaMA-3-8B \citep{llama3}, Gemma-3~\citep{gemmateam2025gemma3technicalreport}. Additionally, we evaluated two closed-sourced LLMs: Gemini-2.5~\citep{geminiteam2025geminifamilyhighlycapable} and GPT-5~\citep{openai2025gpt5}. These models were selected as representative of smaller and resource-efficient architectures. All the specific versions with their respective parameter sizes are listed in Table~\ref{tab:overall_performance}.

\paragraph{Prompt Engineering}
We adopted the GPT-NER methodology \citep{wang2023gpt}, implementing a single-turn per-label prompting strategy, where each advertisement was processed 15 times, once for each entity type. The refined prompting approach incorporated three key components:
\begin{itemize}
    \item \textbf{Task-specific prompts}: Each prompt focused on a single entity type with a concise task description and label definition
    \item \textbf{Few-shot examples}: 2 few-shot in-context learning examples were used.
    \item \textbf{Simplified output format}: The output would copy verbatim the example that we want labelled and the entities of interested would be marked with boundary delimiters (@@...||) rather than full NER formatting.
\end{itemize}

\paragraph{Self-Verification}
Following \citep{wang2023gpt}, a self-verification step was implemented to mitigate the high false positive rate characteristic of LLM-based NER. This secondary prompt presented the model's initial predictions for validation, significantly improving precision whilst marginally reducing recall. Preliminary experiments, reported in Appendix~\ref{sec:app:prelim_verification} (Appendix Table~\ref{tab:prelim_verification}), demonstrated the efficacy of including this step.

Complete prompt templates are provided in Appendix~\ref{sec:appendix:llm-prompts}.

\section{Results}
\label{sec:results}

\subsection{Rules-based Approach}
\label{ssec:rules_based}

The rules-based approach utilised a combination of taxonomy-based string matching and regular expressions. Performance was highly dependent on the complexity of the entity and the matching method employed. Methods involving partial matching via Fuzzy and Levenshtein algorithms, while more accurate for variable text, were significantly slower at inference time compared to exact matching.

Table~\ref{tab:label_performance} summarises the best F1-scores achieved for each label. As shown, simpler, more regular entities like \texttt{YEAR}, \texttt{MAKE}, and \texttt{MODEL} achieved strong performance (0.77--0.79 F1). In contrast, more complex and nuanced labels such as \texttt{INTERIOR\_COLOUR} and \texttt{EXTERIOR\_COLOUR} performed poorly, with F1-scores below 0.03.

An analysis of precision and recall scores reported in Appendix~\ref{app:rules_pr} (Appendix Table~\ref{tab:rules_precision_recall}) reveals that regex-based methods, where applicable (e.g., for \texttt{YEAR}), achieved high recall at the cost of precision, whereas taxonomy-based matching often resulted in higher precision but lower recall, as it failed to capture variations not present in the pre-defined lists. Overall, when evaluated across all present labels, the best-performing configuration of the rules-based approach achieved a micro-averaged F1-score of 0.43.

\subsection{Encoder-based Models}
\label{ssec:encoders}
We evaluated four transformer-based encoders: BERT, RoBERTa, ModernBERT and DeBERTa. Table~\ref{tab:overall_performance} summarises the character-level performance of all evaluated models. Fine-tuned encoder models achieve higher performance, with micro-F1 scores ranging from 82.9\% to 90.1\%, compared to LLMs which achieve only 77.8\% to 41.7\% micro-F1.

\begin{center}
\begin{minipage}{\columnwidth}
\centering
\small
\begin{tabular}{@{}lcc@{}}
\toprule
\textbf{Model} & \textbf{Type} & \textbf{Micro-F1\textdownarrow
} \\
\midrule
microsoft/deberta-v3-base & Encoder & 0.901 \\
FacebookAI/roberta-base & Encoder & 0.895 \\
bert-base-cased & Encoder & 0.873 \\
answerdotai/ModernBERT-base & Encoder & 0.829 \\
\midrule
Gemini-2.5-Flash-Lite & LLM & 0.778 \\
GPT-5-Nano & LLM & 0.775 \\
Gemma-3-27B-IT & LLM & 0.753 \\
Qwen3-30B-A3B-Instruct & LLM & 0.735 \\
Llama-3.1-8B-Instruct & LLM & 0.682 \\
Mistral-7B-Instruct-v0.2 & LLM & 0.417 \\

\bottomrule
\end{tabular}
\captionof{table}{Overall character-level performance comparison. Encoders consistently outperform LLMs with 2-shot in-context learning. The micro-F1 scores are shown in descending order.}
\label{tab:overall_performance}
\end{minipage}
\end{center}

The DeBERTa model demonstrated the best overall performance, achieving an F1 score of 90.1\%. After resampling the data splits and repeating the experiments three times, the standard deviation was 2.3\%, indicating relatively stable performance across runs. This strong performance may be attributable to DeBERTa’s disentangled attention mechanism, which more effectively captures positional information, a feature that is particularly relevant for the consistently structured AI-generated adverts.

Table~\ref{tab:label_performance} shows per-label F1 scores for all the family models. The encoders perform best on numeric entities such as \textbf{YEAR} and \textbf{BATTERY\_CAPACITY}, standardised attributes like \textbf{FUEL\_TYPE} and \textbf{TRANSMISSION}, and vehicle identifiers including \textbf{MAKE} and \textbf{MODEL}. In contrast, they struggle with more subjective attributes such as \textbf{INTERIOR\_COLOUR} and \textbf{EXTERIOR\_COLOUR}, likely due to the wide lexical variation in colour descriptions (e.g., ``midnight blue'', ``pearl white'', ``metallic silver'').

\begin{added}
Part of this residual error appears to be a property of the task rather than of the models. Restricting attention to the ten labels with at least 30 test instances, per-label encoder F1 correlates with the per-label inter-annotator agreement of Appendix Table~\ref{tab:schema} (Spearman $\rho = 0.81$, $p = 0.005$). The clearest case is \texttt{INTERIOR\_COLOUR}, where the best encoder scores .623 against a human agreement of .639: the model is close to the annotation ceiling, and its low score partly reflects a category whose extent annotators cannot themselves fix consistently. By contrast, \texttt{EXTERIOR\_COLOUR} (.725 against an agreement of .866) and \texttt{BODY\_TYPE} (.721 against .867) retain a genuine gap and are more promising targets for improvement.
\end{added}

Our analysis of the training data size indicated that the dataset was sufficient for the task, as model performance (measured by evaluation loss) began to plateau after approximately 300--350 training samples. Further details and the corresponding graph are available in Appendix~\ref{app:training_size}. The one exception is the \texttt{NO\_SEATS} which was scarcely reported in the text descriptions.

\begin{strip}
\centering
\small
\begin{tabular}{@{}lcccccc@{}}
\toprule
\textbf{Entity Type} & \textbf{Best Encoder} & \textbf{Encoder F1} & \textbf{Best LLM} & \textbf{LLM F1} & \textbf{Rules F1} \\
\midrule
MAKE & RoBERTa & \textbf{0.938} &  GPT-5-Nano & 0.825 & 0.773 \\
MODEL & RoBERTa & \textbf{0.950} & Gemini-2.5-Flash-Lite &  0.878 & 0.789 \\
TRIM & DeBERTa-v3 & 0.870 & Gemini-2.5-Flash-Lite & \textbf{0.715} & 0.482 \\
YEAR & DeBERTa-v3 & \textbf{0.982} & Gemini/Gemma & 0.948 & 0.773 \\
BODY\_TYPE & DeBERTa-v3 & 0.721 & GPT-5-Nano & \textbf{0.769} & 0.531 \\
FUEL\_TYPE &  BERT-cased & \textbf{0.957} & Gemini-2.5-Flash-Lite & 0.773 & 0.574 \\
TRANSMISSION & DeBERTa-v3 & \textbf{0.938} & GPT-5-Nano & 0.476  & 0.255 \\
ENGINE\_SPEC & DeBERTa-v3 & \textbf{0.933} & Gemini-2.5-Flash-Lite & 0.907  & 0.279 \\
EXTERIOR\_COLOUR & DeBERTa-v3 & 0.725 & GPT-5-Nano & \textbf{0.763} & 0.024 \\
INTERIOR\_COLOUR & RoBERTa & \textbf{0.623} & Gemini/GPT &  0.270 & 0.007 \\
NO\_SEATS & N/A & 0.000 & Gemma-3-27B-IT  & \textbf{0.750} & 0.258 \\
BOOT\_SIZE & DeBERTa-v3 & 0.909 & Gemini-2.5-Flash-Lite & \textbf{1.000 }& --- \\
BATTERY\_CAPACITY & DeBERTa-v3 & \textbf{1.000 }& Gemini/GPT & 0.714 & 0.462 \\
BATTERY\_RANGE & DeBERTa-v3/RoBERTa & 0.800 & Gemini-2.5-Flash-Lite & \textbf{1.000} & --- \\
RECHARGE\_TIME & RoBERTa & \textbf{1.000} & N/A & 0.000 & --- \\
\bottomrule
\end{tabular}
\captionof{table}{Per-label F1 scores comparing encoder, LLM, and rules-based performance. In bold are the top performants for each label. Some model names are shortened for illustration purposes but are the same version with the ones presented in Table~\ref{tab:overall_performance}.}
\label{tab:label_performance}
\end{strip}

\subsection[LLM-based Models]{\colorbox{addgreen}{LLM-based Models}}
\label{ssec:llms}

The decoder models, which are much larger in parameter size---\del{many times more than 100 times larger}\ins{ranging from tens to hundreds of times more parameters}---than their encoder counterparts excelled in various labels, especially in those with less support, comparatively to the encoders.

Gemini-2.5 scored the highest amongst the LLMs but still second to the encoders. Its predictions across the test set reveal heterogeneous performance across entity types: strong (F1 $>$ 75\%) on YEAR, \texttt{MAKE}, and \texttt{MODEL}; moderate (F1 50--75\%) on \texttt{ENGINE\_SPEC}, \texttt{EXTERIOR\_COLOUR}, \texttt{TRIM}, \texttt{BATTERY\_CAPACITY}, and \texttt{BODY\_TYPE}; and poor (F1 $<$ 50\%) on \texttt{INTERIOR\_COLOUR}, \texttt{TRANSMISSION}, and notably \texttt{RECHARGE\_TIME} (0\% F1 despite 10 test instances).

LLMs exhibit \del{three}\ins{two} primary failure modes. First, \textit{boundary detection errors}: extracting ``300 miles'' instead of ``approximately 300 miles'' \texttt{BATTERY\_RANGE}), or ``408'' instead of ``408 horsepower'' (\texttt{ENGINE\_SPEC}). Second, \textit{variant normalisation failures}: achieving high recall on simple \texttt{FUEL\_TYPE} values (``petrol'', ``diesel'', ``electric'') while missing hybrid variants (``Petrol Hybrid'', ``Petrol Plug-in Hybrid'' , ``Diesel Hybrid'').

We identify three primary factors contributing to these LLM failures. First, our 2-shot learning approach may not be indicative enough for the LLMs to perform inference. Second, vehicle specifications involve domain-specific terminology (``kWh'', ``bhp'', ``T-GDI'', ``BiTurbo'') and complex entity boundaries (e.g., ``2.0L Turbocharged Inline-4'' as \texttt{ENGINE\_SPEC} entities), for which pre-trained LLMs lack adequate exposure during training, unlike general entities such as persons or organisations. Third, the severe lexical variation within certain entity types (e.g., \texttt{INTERIOR\_COLOUR}: ``Bengal red Nappa leather'', ``Light Oyster and Ebony'', ``Full Black Valcona Leather'') exceeds what can be captured in 2-shot demonstrations, while more standardised types (\texttt{YEAR}, \texttt{MAKE}) benefit from prior knowledge in the pre-training corpus.

\del{Albeit,}\ins{That said,} the LLMs proved resourceful, achieving the highest F1 score amongst the other approaches, when the support was quite low in the labels \texttt{NO\_SEATS}, \texttt{BATTERY\_RANGE} and \texttt{BOOT\_SIZE}.

\subsection{Computational Resources}
\label{sec:setup}

All encoder-based transformer models were fine-tuned on an Apple M4 MacBook. Open-source LLM inference was performed using vLLM~\citep{kwon2023efficient} on NVIDIA RTX PRO 6000 (96GB VRAM) or NVIDIA H200 SXM (141GB VRAM) GPUs, depending on model size. Closed-source models were accessed through their respective provider APIs. To ensure deterministic outputs, we set the temperature to 0.0 where supported and used greedy decoding for all generations. Additional hardware and runtime details are provided in Appendix~\ref{app:hardware}.

\section{Discussion}
\label{sec:discussion}
The current investigation has implications for deploying NER systems in production automotive applications. When the goal is extracting entity types and populating structured vehicle databases, fine-tuned encoder models---despite their training requirement---remain the best solution overall, with DeBERTa leading when the supporting examples are sufficient, followed by the LLMs and last the rule-based approaches.

\begin{added}
We expect the three model families to transfer very differently to advertisements from other platforms, countries or writing styles. The rules-based system is the least portable by construction: its taxonomy-matching component is a gazetteer of makes, models and trims supplied by a single UK marketplace, so it would require a new taxonomy for every market it is deployed in, and its regular expressions assume UK conventions such as four-digit registration years and metric capacities. The encoders should degrade more gracefully, since most of what they exploit is local morphology and context rather than a closed vocabulary, but they will still encounter genuinely unseen material: manufacturer-specific colour names (``Santorini Black'', ``Mythos Black'') that behave lexically like arbitrary proper nouns, market-specific abbreviations, and the right-hand-drive UK model line-ups that dominate our \texttt{MODEL} and \texttt{TRIM} distributions. Non-UK listings would additionally introduce vocabulary absent here altogether, and non-English listings fall outside the scope of these models entirely.
\end{added}

Looking beyond immediate deployment constraints and efficiency considerations, several avenues could improve LLM performance on this task. Increasing the few-shot to more examples per entity type could improve prediction accuracy. Similarly, retrieval-augmented prompting could dynamically retrieve relevant examples that could boost the score as well. Finally, models further pre-trained or post-trained on automotive corpora may better handle specialised terminology and entity patterns.

The high accuracy of the fine-tuned encoder models enables several valuable downstream applications. While this work focused on advertisements, the models could be evaluated for tracking vehicles from unstructured social media posts to monitor consumer trends, popular features, and brand perception. Additionally, the process of extracting entities can be used to detect hallucinations in AI-generated content by verifying whether a model that generated the content included additional or different specifications and whether it adhered to the structured vehicle specification data given, thereby enhancing platform integrity. It was notable how on average we could spot an around 20\% increase in F1 score when detecting entities in the AI generated text alone. The fine-tuned encoders could prove resourceful at tracking these mentions from AI generated text, raising flags when necessary.

\section{Conclusions}
\label{sec:conclusions}

In this paper, we addressed the high-impact challenge of extracting fine-grained vehicle specifications from the text descriptions of car advertisements. Our dataset and annotation schema \textbf{AutoSpecNER} comprises 659 advertisements---written by real users and AI-generated proprietary data acquired from the UK's biggest online vehicle-selling website---labelled with 15 specification types ranging from \texttt{MAKE} and \texttt{MODEL} to \texttt{RECHARGE\_TIME}, with the schema validated at a commendable 91.5\% inter-annotator F1 score\ins{, though agreement ranges from .968 for \mbox{\texttt{MAKE}} to .639 for \mbox{\texttt{INTERIOR\_COLOUR}}}. Benchmarking rule-based, fine-tuned encoder, and few-shot decoder approaches, we found the fine-tuned encoders exceptionally well-suited to this task\ins{, in line with the findings of \mbox{\citet{naguib-etal-2024-shot}}}, with DeBERTa achieving a top micro F1-score of 0.901, followed by the decoders (led by Gemini-2.5) and last the rule-based approaches; the LLMs proved particularly resourceful for low-support labels scarcely reported in the text. Finally, this dataset can support market trend analysis and, crucially, the detection of factual hallucinations in AI-generated content.

\section{Limitations}
\label{sec:limitations}

The dataset was drawn exclusively from a proprietary database maintained by a single UK-based company. Although substantial, it is unlikely to capture the full diversity of linguistic usage, document structures, and formatting conventions observed across different countries, platforms, and automotive retailers. In addition, the corpus is UK-specific and monolingual (English), preventing assessment of cross-lingual performance. Improving generalisability would require a more heterogeneous, multi-source, and multilingual dataset.

\begin{added}
\paragraph{Low-support categories.} Five labels have fewer than 20 test instances---\texttt{BATTERY\_RANGE} (3), \texttt{NO\_SEATS} (6), \texttt{RECHARGE\_TIME} (10), \texttt{BOOT\_SIZE} (12) and \texttt{TRANSMISSION} (16), marked in Table~\ref{tab:dataset_splits}---so per-label scores on these are point estimates over very small samples, and differences between models on them are not interpretable. We therefore draw conclusions about model ranking only from the micro-averaged figures. Resampling the splits and repeating the experiments three times gave a standard deviation of 2.3\% in overall micro-F1, comparable to the gaps among the encoders themselves, so we do not claim that the ordering within the encoder family is significant.

\paragraph{Two-shot LLM configuration.} Our LLM results characterise one prompting configuration rather than the general capability of these models. The shot count was fixed at two after a small pilot with two open-weight models using no more than ten demonstrations; this is far too small a fraction of the dataset to establish whether larger demonstration budgets would help. The demonstrations themselves are drawn from this corpus, so the LLM scores benefit from in-domain exemplars that a cold-start deployment would not have. We did not evaluate larger demonstration budgets for the closed-source models, retrieval-augmented exemplar selection, or fine-tuning; the encoder--LLM gap reported here therefore compares fine-tuned encoders with few-shot prompted LLMs specifically, and is not an upper bound on decoder-based approaches.
\end{added}

\section{Ethical Considerations}
The development of the \textbf{AutoSpecNER} dataset and its associated models has been guided by key ethical considerations. The dataset was constructed from publicly accessible advertisements available on a major commercial platform. A manual review of the user-generated content was conducted, which confirmed the absence of Personally Identifiable Information (PII) in the source data. While the data's public availability and freedom from PII mitigate privacy concerns, its origin introduces a risk of socio-economic bias. The model may exhibit performance disparities across content from different seller types, and if the system is more accurate on professionally formatted advertisements than on informal text from private sellers, its application could inadvertently create an unfair marketplace.

Consideration must also be given to the potential for misuse and downstream harms. The same technology that extracts specifications can be repurposed for malicious ends. Models trained on this data could be exploited to generate convincing but fraudulent vehicle advertisements, either for individual scams or at scale to manipulate market perceptions of a vehicle's value and availability. In a deployed application, over-reliance on the system's output could lead to automation bias, where users and platform administrators place undue trust in the automated extractions. This could cause model errors to be systematically propagated, leading consumers to make purchasing decisions based on incorrect information.

Finally, we address the environmental impact of this research. Our experiments involved training and evaluating multiple large-scale models, which consumed significant computational resources and energy. However, our findings make a positive contribution in this area by demonstrating that smaller, fine-tuned encoder models significantly outperform their much larger and more resource-intensive LLM counterparts for this task. This result advocates for a more sustainable approach to deploying NLP in production environments, showing that for specialised industrial tasks, more energy-efficient models can also be the most effective.

\begin{added}
\section*{Acknowledgements}
We thank Autotrader UK for providing access to the underlying advertisement data and for funding the Sponsored Masters Programme at Manchester Metropolitan University under which this work was conducted. We also thank the anonymous reviewers for their constructive and encouraging comments.
\end{added}

\bibliography{main_bib,lr_bib}


\appendix

\section{Preliminary Self-Verification Experiments}
\label{sec:app:prelim_verification}

We conducted preliminary experiments to assess the effect of the LLM self-verification step. Verification improved overall F1 for both Mistral-7B and LLaMA3-8B, mainly by increasing precision while slightly reducing recall. The effect was strongest for Mistral, whose weighted F1 increased from 0.375 to 0.409. For LLaMA3, the gain was smaller, with weighted F1 increasing from 0.473 to 0.481. However, verification produced larger gains for some lower-support labels. For instance, with LLaMA3, the F1-score for \texttt{BATTERY\_CAPACITY} increased from 0.291 to 0.552 and for \texttt{BOOT\_SIZE} from 0.112 to 0.213.

\begin{table}[h!]
\centering
\small
\begin{tabular}{lcc}
\toprule
\textbf{Model} & \textbf{Initial F1} & \textbf{Verified F1} \\
\midrule
Mistral-7B, macro & 0.184 & 0.211 \\
Mistral-7B, weighted & 0.375 & 0.409 \\
LLaMA3-8B, macro & 0.204 & 0.209 \\
LLaMA3-8B, weighted & 0.473 & 0.481 \\
\bottomrule
\end{tabular}
\caption{Effect of self-verification in preliminary LLM experiments.}
\label{tab:prelim_verification}
\end{table}

\section{LLM Prompts}
\label{sec:appendix:llm-prompts}

We use a per-entity extraction strategy following \citet{wang2023gpt}. Each advertisement is processed once per entity type using the prompts below.

\subsection*{Extraction Prompt}

\noindent\textbf{System:}
\begin{quote}
\texttt{You are a NER Model. Your task is to label \{entity\_type\} entities in the given sentence.}
\end{quote}

\noindent\textbf{User:}
\begin{quote}
\texttt{Below are some examples.}\\[0.5em]
\texttt{\{few\_shot\_examples\}Input: \{text\}}\\
\texttt{Output:}
\end{quote}

\noindent where each few-shot example follows the template:
\begin{quote}
\texttt{Input: \{input\_text\}}\\
\texttt{Output: \{output\_text\}}
\end{quote}

\subsection*{Self-Verification Prompt}

\noindent\textbf{System:}
\begin{quote}
\texttt{You are a verification expert. Your task is to verify if an extracted entity is correct.}
\end{quote}

\noindent\textbf{User:}
\begin{quote}
\texttt{Original text:}\\
\texttt{\{text\}}\\[0.5em]
\texttt{Entity type: \{entity\_type\}}\\
\texttt{Extracted entity: \{entity\}}\\[0.5em]
\texttt{Is this extraction correct and accurate? Consider:}\\
\texttt{1. Is the entity actually present in the text?}\\
\texttt{2. Is the entity type classification correct?}\\
\texttt{3. Are the boundaries accurate (no extra or missing characters)?}\\[0.5em]
\texttt{Answer only with a ``yes'' if all statements are true, otherwise say ``no''. Do not say anything more.}\\[0.5em]
\texttt{Verification result:}
\end{quote}

\section{Rules-based Precision and Recall}
\label{app:rules_pr}
Table~\ref{tab:rules_precision_recall} show the per-label precision and recall scores for the best-performing rules-based methods.

\begin{table}[h!]
\centering
\caption{Precision and Recall Scores for Best Rules-Based Methods.}
\label{tab:rules_precision_recall}
\begin{tabular}{lcc}
\toprule
\textbf{Label} & \textbf{Precision} & \textbf{Recall} \\
\midrule
battery          & 1.000 & 0.300 \\
model            & 0.880 & 0.775 \\
body\_type       & 0.824 & 0.500 \\
make             & 0.666 & 0.949 \\
year             & 0.631 & 0.997 \\
trim             & 0.575 & 0.584 \\
fuel\_type       & 0.479 & 0.717 \\
engine\_spec     & 0.295 & 0.289 \\
transmission     & 0.197 & 0.655 \\
number\_of\_seats & 0.154 & 0.800 \\
exterior\_colour & 0.015 & 0.134 \\
interior\_colour & 0.004 & 0.064 \\
\bottomrule
\end{tabular}
\end{table}

\section{Encoder Training Set Size Analysis}
\label{app:training_size}
\begin{figure*}[h!]
\centering
\includegraphics[width=0.8\textwidth]{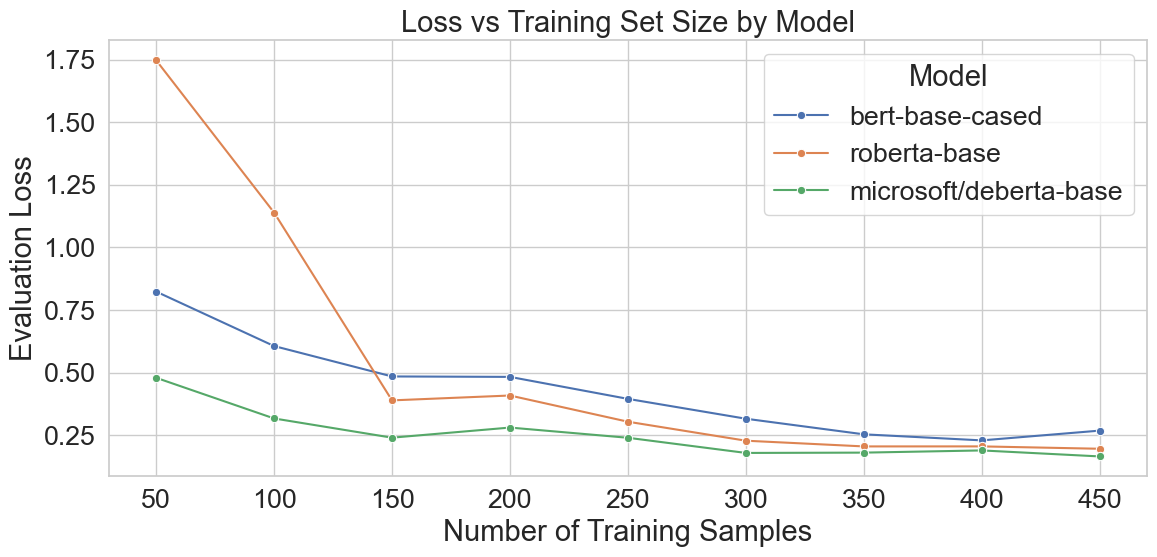} 
\caption{Evaluation loss vs. number of training samples for encoder models.}
\label{fig:loss_vs_samples}
\end{figure*}

Figure~\ref{fig:loss_vs_samples} shows the evaluation loss of the encoder models relative to the number of training samples. The plateauing of the loss curve suggests that the dataset size of 659 adverts was sufficient for the models to learn the primary patterns in the data\ins{ in aggregate; as noted in \mbox{Section~\ref{sec:limitations}}, this does not hold uniformly across labels, and the sparsest categories remain under-supported}.

\section{Hardware and Runtime Details}
\label{app:hardware}

All encoder-based transformer models were fine-tuned on an Apple M4 MacBook Pro equipped with 36GB unified memory. Inference for the encoder-based models required between a few seconds and approximately one minute to process the complete test set.

For open-source LLMs, models with fewer than 10 billion parameters were executed using vLLM~\citep{kwon2023efficient} on an NVIDIA RTX PRO 6000 GPU with 96GB VRAM. Larger models were executed on an NVIDIA H200 SXM GPU with 141GB VRAM.

LLM inference included both the entity extraction and entity verification stages. Processing the full evaluation set required approximately 1.5 hours per model. This corresponded to approximately $109 \times 15 \times 2 = 3,270$ model queries, where 109 is the number of test samples, 15 is the number of target entities, and 2 corresponds to the verification stage.

Closed-source models (e.g., GPT and Gemini) were accessed through their respective provider APIs. All LLMs were interfaced using an OpenAI-compatible API format. To ensure deterministic outputs, we set the temperature to 0.0 where supported by the API and used greedy decoding for all generations.

\end{document}